\documentclass[
]{ceurart}

\usepackage{listings}
\usepackage{url}
\usepackage{color}
\usepackage{graphicx}

\begin{document}

\copyrightyear{2022}
\copyrightclause{Copyright for this paper by its authors.
  Use permitted under Creative Commons License Attribution 4.0
  International (CC BY 4.0).}

\conference{De-Factify 2: Second Workshop on Multimodal Fact Checking and Hate Speech Detection, co-located with AAAI 2023}

\title{NYCU-TWO at Memotion 3: Good Foundation, Good Teacher, then you have Good Meme Analysis}


\author[1]{Yu-Chien Tang$^{\dagger}$}[%
email=tommytyc.cs10@nycu.edu.tw,
]

\author[1]{Kuang-Da Wang$^{\dagger}$}[%
email=gdwang.cs10@nycu.edu.tw, 
]
\author[1]{Ting-Yun Ou$^{\dagger}$}[%
email=outingyun.cs11@nycu.edu.tw 
]
\author[1]{Wen-Chih Peng}[%
email=wcpeng@nctu.edu.tw,
]

\address[1]{National Yang Ming Chiao Tung University, Hsinchu, Taiwan}

\def\thefootnote{$\dagger$}\footnotetext{Equal contributions.}\def\thefootnote{\arabic{footnote}}

\begin{abstract}
This paper presents a robust solution to the \textit{Memotion 3.0 Shared Task}. The goal of this task is to classify the emotion and the corresponding intensity expressed by memes, which are usually in the form of images with short captions on social media. Understanding the multi-modal features of the given memes will be the key to solving the task. In this work, we use CLIP\cite{radford2021learning} to extract aligned image-text features and propose a novel meme sentiment analysis framework, consisting of a Cooperative Teaching Model (CTM) for Task A and a Cascaded Emotion Classifier (CEC) for Tasks B\&C. CTM is based on the idea of knowledge distillation, and can better predict the sentiment of a given meme in Task A; CEC can leverage the emotion intensity suggestion from the prediction of Task C to classify the emotion more precisely in Task B. Experiments show that we achieved the 2nd place ranking for both Task A and Task B and the 4th place ranking for Task C, with weighted F1-scores of 0.342, 0.784, and 0.535 respectively. The results show the robustness and effectiveness of our framework. Our code is released at github\footnote{https://github.com/tommytyc/Memotion-DeFactify-AAAI-2023}.
\end{abstract}


\maketitle

\section{Introduction}
There are two common definitions\cite{Meme_meaning} of a meme: (1) an amusing or interesting item (such as a captioned picture or video) or genre of items which spread widely online, especially through social media; (2) an idea, behavior, style, or usage that spreads from person to person within a culture. With careful analysis of the underlying sentiment of a widespread meme, people can get a better understanding of the post content from social media. However, due to the multi-modal nature of the meme, it is no easy task to understand its emotion and intensity only with the image content or its caption, hindering the potential application, such as detecting hateful or harmful memes.
Considering the strong correlation between the images and captions\cite{hu2018multimodal}, downstream emotion classification tasks and sentiment analysis can benefit from high-quality multi-modal representation. We take advantage of the CLIP\cite{radford2021learning} model, which is pre-trained with contrastive loss and is able to align the multi-modal features in high-dimensional embedding space, as a foundation to retrieve the rich information inside images and text. 

Besides, we observe that the sentiment label and its scales are hierarchical (e.g., the emotion \textit{humorous} contains \textit{funny}, \textit{very funny}, \textit{hilarious} in Task C), and thus introduce two different models, CTM and CEC, for the different downstream tasks. In Task A, we observe that different types of sentiment are composed of different proportions of positive and negative emotions. Therefore, we propose CTM, which introduces the concept of knowledge distillation and uses the framework of the teacher-student model. The good teacher and the bad teacher will cooperate with each other and teach their own students to achieve better performance on Task A. CEC considers the hierarchical characteristics of emotions in the model architecture, predicts the emotion intensity for Task C, and leverages the prediction as a suggestion to classify the emotion for Task B so that both Task B can achieve better performance, compared to using a single model.

\section{Related Work}

\textbf{Meme Understanding.} People express themselves with memes in various templates on social media as a way of communication. Modern memes are images with an embedded short text. While sentiment analysis in memes needs to extract features from both modalities, some researchers adopt multi-modal deep neural networks to analyze the sentiment of memes. In previous competitions, many different deep learning approaches have been developed, such as multi-task classification networks and multi-modal models \cite{lee2020amazon, bucur2022blue}. Previous studies usually adopt fusion techniques to aggregate features from text and images to obtain multi-modal information for better sentiment classification performance\cite{tensoremnlp17, tsai2019MULT}, but none of them has shed light on the hierarchical features of sentiment labels.

\textbf{Vision-Language Pre-training.} Recently there have been plenty of multi-modal models combining modules from different fields in various design ways. They have had surprising results, especially in the image-text field. ConVIRT\cite{zhang2020contrastive} uses paired descriptive text to learn medical visual representations successfully, while CLIP\cite{radford2021learning} has impressive performance on the zero-shot transfer model to downstream tasks by pre-training huge amounts of image-text pairs data and modifying the ConvIRT\cite{zhang2020contrastive} architecture. The Google research team proposed CoCa\cite{yu2022coca}, an image-text encoder-decoder foundation model pre-trained with contrastive loss and captioning loss. It has the ability of contrastive approaches like CLIP\cite{radford2021learning} and generative methods like SimVLM\cite{wang2021simvlm}. In this challenge, we use CLIP as a multi-modal feature encoder to extract rich vision-language information from the meme.

\textbf{Knowledge Distillation.} Knowledge distillation is a technique used in model compression\cite{Bucila2006ModelC, hinton2015distilling}. The main concept is to extract the knowledge from a complex model for another simple model so that this small simple model can also achieve the same effect as the complex model. In the vanilla setting, it is usually implemented in the framework of the teacher-student concept: a large deep neural network is regarded as a teacher training a smaller student neural network from its logits. Even when the teacher model and student model are the same, it can still improve the generalization and robustness of semi-supervised models. The framework with the same architecture as the teacher model and student model is called self-distillation\cite{zhang2019your}. The Cooperative Teaching Model (section 4.2) is based on self-distillation and provides the teacher with more additional information to make it easier to learn.

\section{Task Description}
The Memotion 3.0\cite{mishra2023memotionoverview} shared task is the third iteration of the Memotion task which was first conducted at Semeval 2020. The Memotion 3.0 \cite{mishra2023memotion3} dataset is made up of training dataset, validation dataset, and testing dataset at the ratio of 5:1:1. Each sample includes an image and the corresponding captions extracted by the OCR system. In Table \ref{tab:title}-\ref{tab:title3}, we show the details and the label distributions for each of the different tasks:

\begin{itemize}
\item \textbf{Task A: Sentiment analysis.} Given a meme image and its caption, the goal is to classify the sentiment into three labels, namely \textit{positive}, \textit{neutral}, and \textit{negative}.
\item \textbf{Task B: Emotion classification.} Given a meme image and its caption, the task aims to identify the types of emotion the meme belongs to, including \textit{humorous}, \textit{sarcastic}, \textit{offensive}, and \textit{motivational}. Each meme can express more than one emotion.
\item \textbf{Task C: Scales/Intensity of Emotion Classes.} The goal of this task is to quantify the intensity of each emotion. The scales of each emotion class are from 0 to 3 for \textit{humorous}, \textit{sarcastic}, and \textit{offensive}, but only 0 and 1 for \textit{motivational}.
\end{itemize}

\begin{table}[h]





\begin{minipage}{\linewidth}
  \centering
    \begin{tabular}{cccccccccc} 
    \toprule 
        \multicolumn{1}{c}{\multirow{2}{*}{dataset}} &\multicolumn{3}{c}{train} & \multicolumn{3}{c}{valid} & \multicolumn{3}{c}{test} \\ 
        \cmidrule(lr){2-4}\cmidrule(lr){5-7}\cmidrule(lr){8-10}
        \centering
          & Neg & Neut & Pos & Neg & Neut & Pos & Neg & Neut & Pos \\
        \midrule 
        \textbf{overall} & 25\%(17:83) & 42\% & 33\%(16:84) & 39\%(18:82) & 38\% & 23\%(11:89) & 39\% & 36\% & 25\% \\

    \bottomrule 
    \end{tabular}
    \caption{The label proportion in Task A. Ratio of two extra labels \textit{very neg} and \textit{very pos} are shown in parentheses. \\ }
    \label{tab:title}
\end{minipage}

\begin{minipage}{\linewidth}
    \centering
    \begin{tabular}{ccccccc}
    \toprule 
        \multicolumn{1}{c}{\multirow{2}{*}{dataset}} &\multicolumn{2}{c}{train} & \multicolumn{2}{c}{valid} & \multicolumn{2}{c}{test} \\ 
        \cmidrule(lr){2-3}\cmidrule(lr){4-5}\cmidrule(lr){6-7}
        \centering
          & Negative& Positive & Negative & Positive & Negative & Positive  \\
        \midrule 
        \textbf{humorous} & 14\% & 86\% & 7\% & 93\% & 7\% & 93\% \\
        \textbf{sarcastic} & 21\% & 79\% & 8\% & 92\%  & 9\% & 91\% \\
        \textbf{offensive}  & 61\% & 39\% & 43\% &  57\%  & 45\% & 55\% \\
        \textbf{motivational}  & 88\% & 12\% & 97\% & 3\% & 96\% & 4\% \\

    \bottomrule 
    \end{tabular}
    \caption{The label proportion in Task B. \\ }
    \label{tab:title2}
\end{minipage}
\begin{minipage}{\linewidth}
    \centering
    \begin{tabular}{ccccccccccccc}
    \toprule 
        \multicolumn{1}{c}{\multirow{2}{*}{dataset}} &\multicolumn{4}{c}{train} & \multicolumn{4}{c}{valid} & \multicolumn{4}{c}{test} \\ 
        \cmidrule(lr){2-5}\cmidrule(lr){6-9}\cmidrule(lr){10-13}
        \centering
          & Not & Little & Very & Extr  & Not & Little & Very & Extr  & Not & Little & Very & Extr  \\
        \midrule 
        \textbf{humorous} & 15\% & 48\% & 29\% & 8\% & 7\% & 65\% & 25\% & 3\% & 7\% & 62\% & 27\% & 4\% \\
        \textbf{sarcastic} & 21\% & 28\% & 43\% & 8\% & 8\% & 65\% & 25\% & 2\% & 9\% & 62\% & 27\% & 2\% \\
        \textbf{offensive} & 61\% & 27\% & 9\% & 3\% & 43\% & 53\% & 3\% & 1\% & 45\% & 51\% & 3\% & 1\% \\
        \textbf{motivational} & 88\% & - & - & 12\% & 97\% & - & - & 3\% & 96\% & - & - & 4\% \\
    \bottomrule 
    \end{tabular}
    \caption{The label proportion in Task C. Extr = Extremely.}
    \label{tab:title3}
\end{minipage}

\end{table}

\section{Methodologies}
\subsection{Meme Encoder}
Several powerful methods\cite{tan2019efficientnet, DBLP:conf/naacl/DevlinCLT19, DBLP:conf/iclr/LanCGGSS20} have been proposed for feature extraction in the vision and language domains. We decided to use two types of encoders to obtain better semantic features for the multi-modal problems: (1) direct features from a Swin Transformer\cite{liu2021swin} which is pre-trained on the ImageNet-21k dataset, and will then be fine-tuned on the Memotion task dataset, and (2) a CLIP\cite{radford2021learning} model. CLIP is composed of an image encoder and a text encoder, both jointly pre-trained to project the image and the caption onto the same embedding space in a contrastive manner. In this way, the extracted image embeddings and the caption embeddings are aligned, and the images will be near the captions with similar semantic features. We adopt ViT\cite{dosovitskiy2021an} as the image encoder and DistilBERT\cite{Sanh2019DistilBERTAD} as the text encoder in our CLIP model.

\textbf{Feature Extraction Pipeline.} For each of the following downstream tasks, the first step of computation is to extract the features of the meme images and their captions. The Swin Transformer and the CLIP image encoder will encode the meme images into two vectors respectively, and the CLIP text encoder will also be used to generate the caption embeddings. The output multi-modal embedding tuple is made up of the above three embeddings.

\begin{figure}
    \centering
    \includegraphics[width=\linewidth]{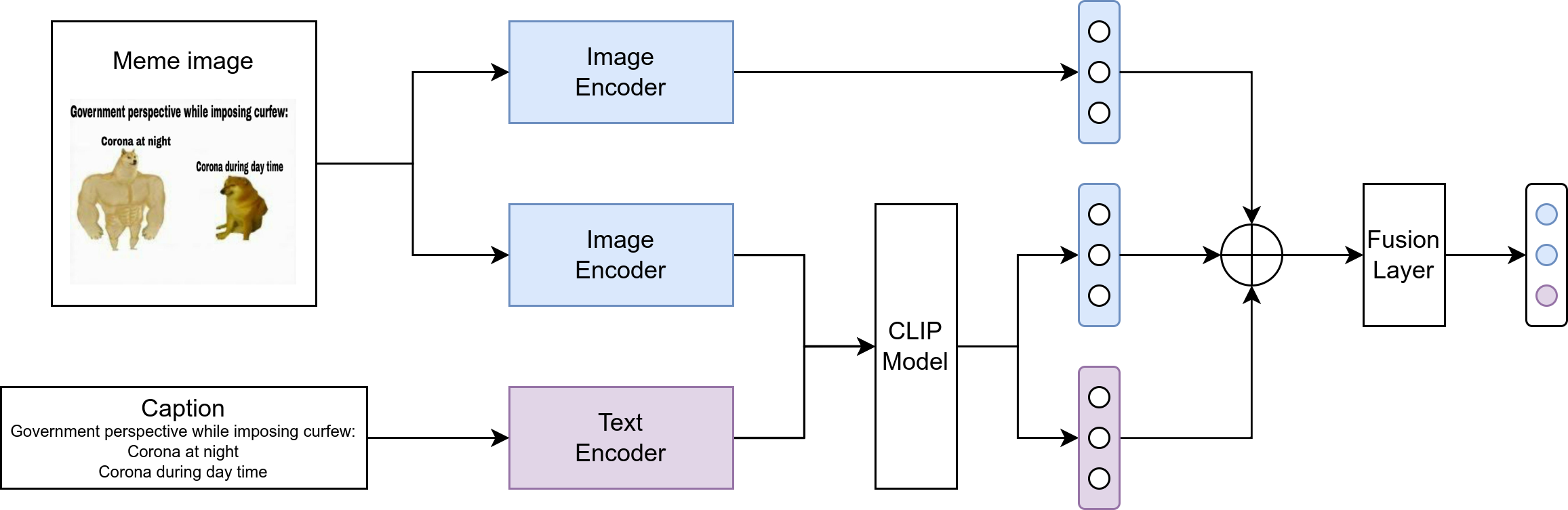}
    \caption{Illustration of the Meme Encoder.}
    \label{fig:encoder}
\end{figure}

\subsection{Task A: Cooperative Teaching Model (CTM)}
\begin{figure}
    \centering
    \includegraphics[width=\linewidth]{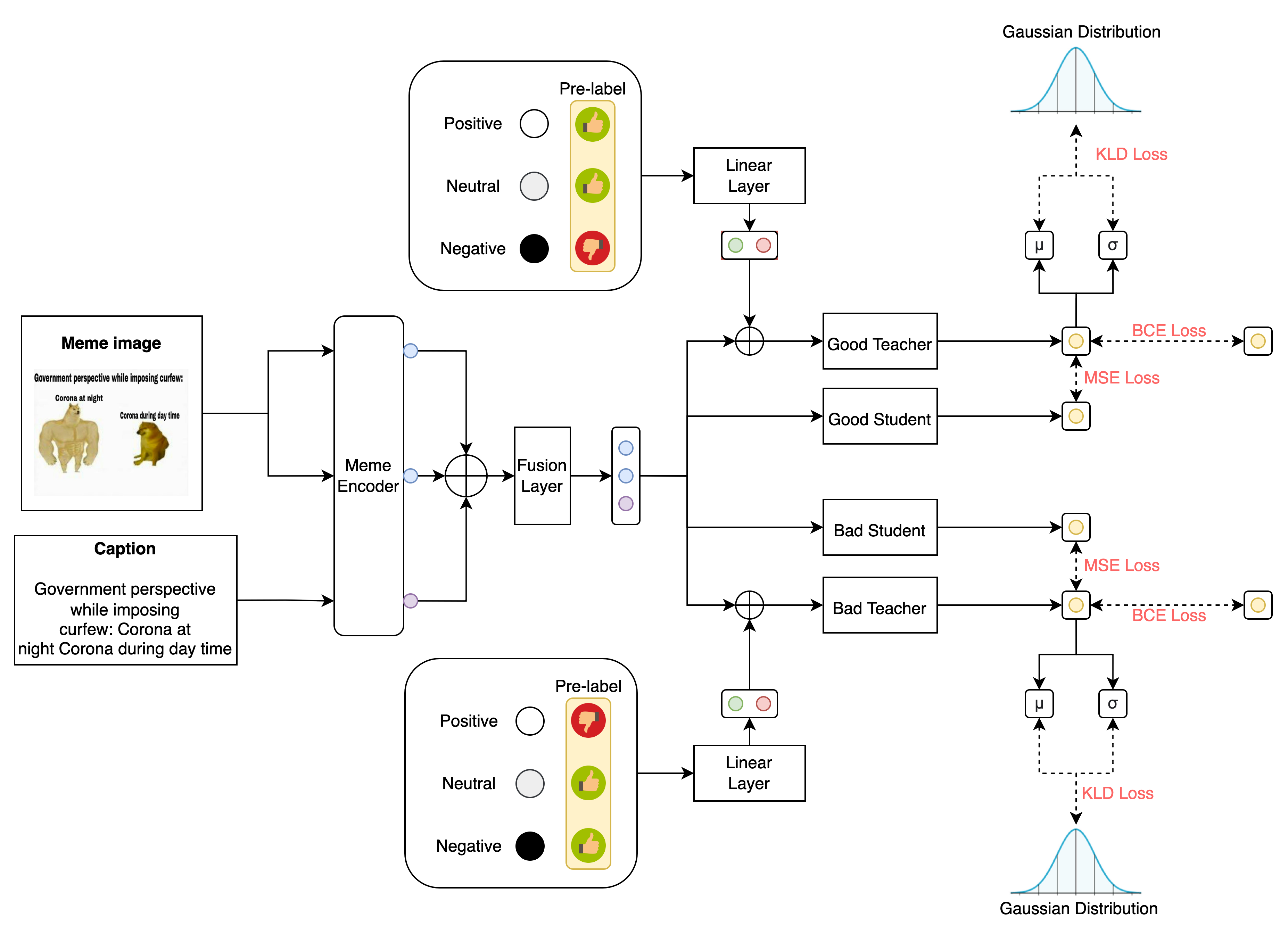}
    \caption{Illustration of the Cooperative Teaching Model (CTM).}
    \label{fig:CTM}
\end{figure}

We present our proposed model for Task A, called the Cooperative Teaching Model (CTM). An overview of the CTM is illustrated in Figure \ref{fig:CTM}. Task A aims to classify the meme into three categories based on the expressed sentiment. However, we believe that the three categories should be regarded as different extents between positive and negative sentiment. That is, the \textit{neutral} actually belongs to either the positive or the negative, but implicitly. Based on this idea, we introduce the concept of knowledge distillation to design the framework that has two \textbf{teacher} models to teach their \textbf{student} models how to classify sentiment respectively. The two teachers are a good teacher and a bad teacher. In the training period, the good teacher teaches students how to judge the positive sentiment of memes, and vice versa. In the inference period, we classify the meme into three classes according to the judgment of the student model.

\subsubsection{Teacher Model}
The difference between the teacher model and the student model is that in addition to the features of the meme images and their captions, the input of the teacher model also includes additional information to help meme sentiment classification. The reason is to make the teacher model worthy of being learned by the student model and to let the teacher model learn faster than the student model.

Since the \textit{neutral} class actually has slight positive or negative sentiment, we regard it as representing both positive and negative sentiment and merge the three categories into two (the pre-label in figure \ref{fig:CTM}). This pre-label will be provided as additional information of input to the teacher model for training, helping the teacher model classify memes more easily.

The goal of the teacher model is to learn how to classify whether the sentiment of the meme is positive or negative, and the results are provided for students to learn. We added a regularization term for the teacher model about the degree of positive and negative sentiment that should conform to a Gaussian distribution. Table \ref{tab:title} shows that the probability of extreme sentiment should be small. Therefore, the output probability distribution of the two teachers should also approach the Gaussian distribution, which will be more realistic.

\subsubsection{Student Model}
The goal of the student model is to approximate the output of the teacher model as much as possible. During the training process of the student model, we record their confidence in the sentiment classification. Just like a real student in the learning process, as long as there is a slight change in a difficult or unread question, it will increase the uncertainty of the student's answer. We bring the learning process of students into the student model and add Gaussian noise to the same meme embedding for disturbance. If the standard deviation of the distribution is small, it can be considered that the student has great confidence in the judgment. In the same way, if the standard deviation of the distribution is large, it can be considered that the students have no confidence in the judgment. Therefore, we train the student models to predict with great confidence by minimizing the standard deviation. We also record the mean of the student models’ prediction of the disturbed meme during the training phase as the threshold for determining whether the meme is \textit{negative} or \textit{positive} during the inference phase. Compared with the general default threshold of 0.5, such a threshold can make the student model have stricter standards for classification and ensure a certain amount of \textit{neutral} predictions.

\subsubsection{Loss function}
We let $N$ be the number of samples. The ground truth is represented by a pre-label during training, so there are only two categories of sentiment, namely positive and negative. We train the Cooperative Teaching Model with the loss function:
\begin{equation}
L = L_{t} + L_{dst} + L_{s} + L_{cfd}
\end{equation}

\begin{itemize}
\item $L_{t}$ is the binary cross-entropy loss of predictions from the teacher model and its corresponding pre-labels.
\begin{equation}
L_{t} = - \frac{1}{N} \sum_{i=1}^{N} (y_{i}\log(p_{i}) + (1-y_{i})\log(1-p_{i})) 
\end{equation}

\item $L_{dst}$ is the Kullback–Leibler divergence between the probability distribution of the teacher model (denoted by $P$) and a Gaussian distribution with learnable mean and variance $N(\mu, \sigma^2)$. It is used to regularize the teacher models to output a more realistic distribution.
\begin{equation}
L_{dst} = D_{kl}(P||N(\mu, \sigma^2))
\end{equation}

\item $L_{s}$ is the mean square error (MSE) between each prediction of the student model and the prediction of the corresponding teacher model.
\begin{equation}
L_{s} = \frac{1}{N}\sum_{i=1}^{N}(p_{i}^{student} - p_{i}^{teacher})^2
\end{equation}

\item $L_{cfd}$ is the standard deviation of the probability distribution from the student model for the same meme with different Gaussian noises; the smaller the standard deviation, the greater the confidence. For each meme, we generate $k$ different meme embeddings with Gaussian noise, where $k$ = 1000 by default.
\begin{equation}
L_{cfd} = \frac{1}{N}\sum_{i=1}^{N}Std(p_{i}^{0},...,p_{i}^{k-1})
\end{equation}
\end{itemize}

\subsection{Tasks B\&C: Cascaded Emotion Classifier (CEC)}
\begin{figure}
    \centering
    \includegraphics[width=\linewidth]{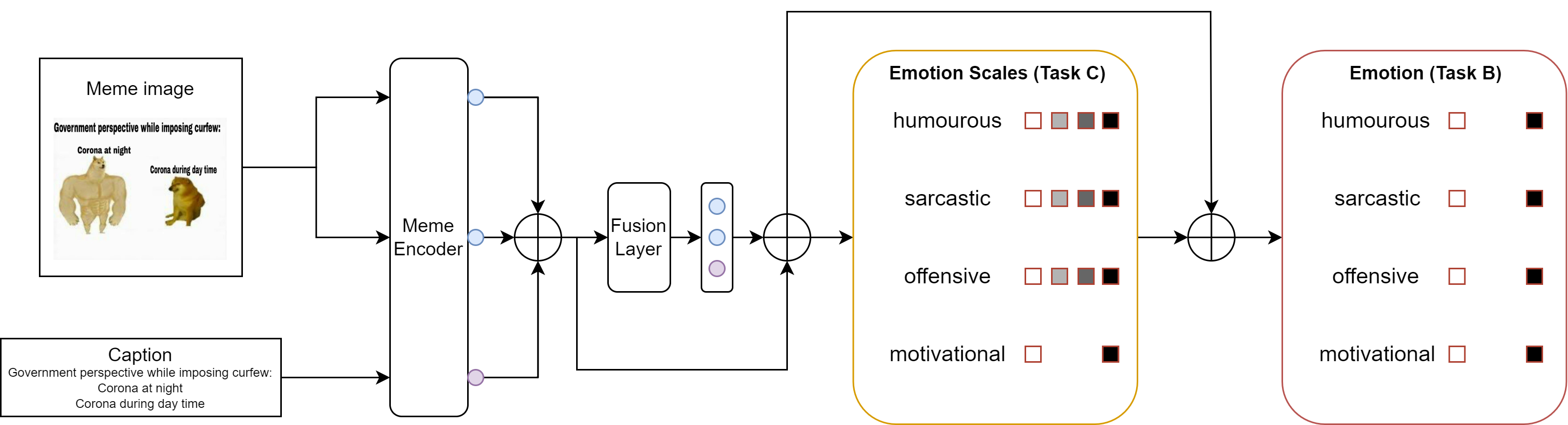}
    \caption{Illustration of the Cascaded Emotion Classifier (CEC).}
    \label{fig:CEC}
\end{figure}
Tasks B and C are essentially related since we can get the prediction of Task B by a simple transformation based on the prediction of Task C. For instance, if the classifier predicts \textit{very offensive} in Task C, the prediction of class \textit{offensive} in Task B can be 1. In the light of this, we propose a framework combining the two classification tasks by leveraging the prediction of Task C as a suggestion for Task B. Specifically, given a meme image and its caption in Task C, a fusion layer will first combine the multi-modal information extracted by the Meme Encoder and generate a fusing embedding. Then the fusing embedding is fed to four MLPs with the multi-modal embedding to predict the corresponding scales for each emotion class. Task B, as an extension of Task C here, will dynamically assess whether the scale prediction of Task C is trustworthy. More precisely, the prediction output of Task C will be concatenated with the multi-modal embedding and be fed to an MLP classifier to predict the emotion expressed by the meme.

\subsubsection{Loss function}
We optimize Tasks B and C with binary cross-entropy loss $L_{B}$ and softmax cross-entropy loss $L_{C}$ respectively, and the total loss is the sum of them. It is worth noting that we simplify the notation with a single loss term for each emotion class.
\begin{equation}
L = L_{B} + L_{C}
\end{equation}
\begin{equation}
L_{B} = - \frac{1}{N}\sum_{i=1}^{N} (y_{i}\log(p_{i}) + (1-y_{i})\log(1-p_{i}))
\end{equation}
\begin{equation}\label{C}
L_{C} = -\frac{1}{N}\sum_{i=1}^{N}log\frac{exp(p_{i,y_{i}})}{\sum_{j \in [K]}exp(p_{i,j})}
\end{equation}
For \eqref{C}, $K$ denotes the number of scales of each emotion class, and $p_{i,j}$ denotes the predicting probability of j-th scale for a sample $i$.

\section{Experiment and Discussion}
For the CLIP model, we pre-train it on three datasets, namely MET-Meme\cite{xu2022sigir}, Memotion 1.0\cite{chhavi2020memotion}, and Memotion 3.0\cite{mishra2023memotion3}. The Memotion 2.0 dataset\cite{ramamoorthy2022memotion} was not available on the Internet so we didn't refer to it. The pre-trained CLIP model is frozen and is not fine-tuned in the downstream tasks. In contrast, the Swin Transformer is fine-tuned in the downstream tasks, as we believe that it can capture different perspectives of features from the CLIP model. All of our experiments were conducted on a machine with an Nvidia GTX 3060 12GB GPU. For Task A, since \textit{neutral} is implicit positive sentiment or negative sentiment, \textit{neutral} will appear only when the predictions of a good student and a bad student are both smaller than each other's threshold. However, during the inference phase, most of the bad student predictions cannot reach the threshold, resulting in many negative sentiment memes being recognized as \textit{neutral}. To correctly classify the \textit{negative} hidden in the \textit{neutral}, we add a judgmental statement in the inference phase: when the prediction of the bad student is greater than the prediction of the good student, the meme is classified as \textit{negative}.

\subsection{Competition Results}
\begin{table*}[h]  
  \begin{tabular}{c|c|c|c}
        \toprule
        Task A & Task B & Task C & Team \\
        \midrule
        \underline{0.342} & \underline{0.783} & 0.535 & NYCU-TWO\\
        \midrule
         0.337 & 0.743  & 0.530 & CUFE\\
         0.332 & 0.747 & 0.522 & Baseline\\
         0.328 & \textbf{0.797} & \textbf{0.598} & wentaorub\\
        \textbf{0.344} & 0.676 & \underline{0.570} & NUAA-QMUL-AIIT\\
        0.333 & 0.722 & 0.539 & CSECU-DSG\\
      \bottomrule
    \end{tabular}
  \centering
  \caption[c]{The competition result. The best result in each column is in boldface while the second best is underlined.}
  \label{tab:competition}
\end{table*}

Table \ref{tab:competition} summarizes our competition results of the 3 tasks. Among the best weighted F1-scores of the three subtasks, we achieved a score of 0.342 for sentiment analysis, 0.783 for emotion classification, and 0.535 for scale/intensity of emotion classes, respectively. We also did a further analysis of the meme data and found two possibilities that can make our model perform better.
\begin{itemize}
\item The text in the Memotion 3.0 is in Hinglish, which affected the performance of the foundation model pre-trained on English data. If we could pre-train CLIP with other Hinglish meme datasets, or if the task was in English, the performance may be improved.
\item The CLIP model can make the images near the captions with similar semantic features by aligning the extracted image embedding and the caption embedding. However, the text in a meme does not simply describe the things in the meme image but has implicit meanings. This means that to correctly classify the sentiment and emotion of a meme, besides recognizing the object or event in the meme image, we need to have enough understanding of culture and society to understand the implicit meaning of the meme with the help of the caption.
\end{itemize}

\subsection{Ablation studies}
An extensive ablation study was conducted to verify the design of the Cooperative Teaching Model (CTM) and the Cascaded Emotion Classifier (CEC). The ablation study for the Meme Encoder was not conducted as it provided the multi-modal embeddings for each downstream task. For CTM, we developed four variants to investigate the relative contributions of different components:  1) w/o TR, which is CTM without the teacher model, and only uses the student model with pre-label to training; 2) w/o TD, which is the student model of the CTM using the default threshold of 0.5 for judging positive or negative during evaluation. We also implement a simple classifier, instead of using a pre-label, connecting the features extracted from the Meme Encoder to a linear layer to classify 3 categories (denoted by a simple classifier). For CEC, we remove the cascaded architecture to analyze the contributions (denoted by w/o C). The performance of all variant models is reported in Table \ref{tab:ablation}. We summarize the observations as follows.

\begin{table*}[h]
  \begin{tabular}{c|c|c}
        \toprule
        Task & Model & Weighted F1 \\
        \midrule
         \multirow{6}{*}{Task A}& w/o TR & 0.3484 \\
         & w/o TD & 0.3491 \\
         & w/o TR \& w/o TD & 0.1029 \\
         & simple classifier & 0.2366 \\
         \cmidrule{2-3}
         & \multirow{2}{*}{CTM} & 0.4774 (Memotion 1.0)\\
       & & 0.3689 (Memotion 3.0) \\
       \midrule
       \multirow{2}{*}{Task B}& w/o C & 0.7885 \\
         & CEC & 0.8126 \\
         \midrule
       \multirow{2}{*}{Task C}& w/o C & 0.6048 \\
         & CEC & 0.5304 \\
      \bottomrule
    \end{tabular}
  \caption[c]{Ablation study of our models on the validation dataset.}
  \label{tab:ablation}
\end{table*}

We observe that all the designs in the CTM and CEC contribute to the corresponding tasks. For CTM, the teacher model and the student model with learned thresholds need to cooperate with each other to further improve the performance. In addition, without both of them will cause a performance decline of 26.6\%, which is 13.37\% lower than the simple classifier. This indicates that the design of merging the three categories into a binary pre-label needs to cooperate with the teacher model and the student model with learned thresholds, and can greatly improve the performance by about 13.23\% more than the performance of the simple classifier. Finally, as mentioned earlier, the text in the Memotion 3.0\cite{mishra2023memotion3} dataset is Hinglish. If we use the same language for pre-training, we may be able to improve the performance. However we were not able to find another Hinglish dataset for more appropriate pre-training, and so decided to use the Memotion 1.0\cite{chhavi2020memotion} dataset for verification. The experimental results show that our method indeed improved performance, reaching a weighted F1-score of 0.4774.

For the CEC, the results in Table \ref{tab:ablation} illustrate that task-specific networks still outperform our model cascading Task B and Task C. However, we believe that the CEC architecture can be a reference for similar emotion classification tasks.

\section{Conclusions \& Future Work}
This work presents Team NYCU\_TWO’s approach to classifying the emotion and the corresponding intensity of memes from social media. Besides a powerful multi-modal feature extraction pipeline with the integration of CLIP, our framework incorporates two models, namely the Cooperative Teaching Model and the Cascaded Emotion Classifier, for Task A and Tasks B\&C. We achieved competitive performance at the end of the challenge, showing the effectiveness of the framework.

For our future work, we plan to improve the model from two different directions. The first one could be the low-resource Hinglish problem, since the pre-trained language model is not trained on Hinglish data as much as it is on English data, and the extracted caption embeddings cannot fully reflect the rich semantic information, including sentiment. Aggregating state-of-the-art methods\cite{wang2020extending, Ogueji2021SmallDN} for low-resource language may be able to address the issue. The second one is the aligning problem of the CLIP model in memes. We find that unlike the common image-text dataset for the VQA problem, in which the text can describe the image well, the captions are not supplementary to meme images. The CLIP model can pull the image and text with similar semantic meaning closer, but this is not the case in the meme image-text pairs here. It will be an interesting survey topic to design a better contrastive learning objective toward meme image-text pre-training.

\bibliography{sample-ceur}




\end{document}